\documentclass{article}
\PassOptionsToPackage{numbers, compress}{natbib}

\usepackage[preprint]{neurips_2026}
\usepackage[]{custom}

\title{\modelname{}: Learning Local Self-Supervised Features for CryoEM Volumes via Hypernetworks}

\author{
  Phillip Lo\thanks{co-first authorship} \\
  Biohub \\
  Chicago, IL \\
  \texttt{phillip.lo@biohub.org} \\
  \And
  Sudarshan Babu\textsuperscript{*} \\
  Biohub \\
  Chicago, IL \\
  \texttt{sudarshan.babu@biohub.org} \\
  \And
  Dari Kimanius\thanks{co-senior authorship} \\
  Biohub \\
  Redwood City, CA \\
  \texttt{dari.kimanius@biohub.org} \\
  \And
  Aly A. Khan\textsuperscript{\textdagger} \\
  University of Chicago, Biohub \\
  Chicago, IL \\
  \texttt{aakhan@uchicago.edu} \\
}

\begin{document}

\maketitle

\begin{abstract}
CryoEM map interpretation requires features that are spatially localized, consistent across samples, and informative across spatial scales. Most deep learning methods for map annotation extract features from fixed voxel grids. However, implicit neural representations (INRs) are able to model volumetric data as scale-agnostic, coordinate-conditioned functions. INRs are therefore attractive for cryoEM, but fitting a separate INR for each map is too expensive for large-scale feature extraction and produces representations that are not aligned across samples. We introduce \modelname{}, a self-supervised framework that amortizes INR fitting for reconstructed cryoEM maps. Pretrained on 5,439 Electron Microscopy Data Bank maps, \modelname{} is a transformer-based hypernetwork that generates high-fidelity reconstructions across a wide range of protein structures, including large multi-subunit assemblies. Beyond reconstruction, the INR generated by the pretrained transformer exposes a continuous, local feature field through its intermediate activations at any spatial query point, a property that voxel grid and patch-tokenizer architectures do not naturally provide. Used as auxiliary channels to a 3D nested U-Net annotation head trained from scratch, these coordinate-conditioned features improve performance on eight voxel-level property prediction tasks over a volume-only baseline. Our results demonstrate that amortized implicit neural representations are an effective primitive for geometry-aware analysis of cryoEM data.
\end{abstract}

\section{Introduction}
\label{sec:introduction}

Cryo-electron microscopy (cryoEM) has become a central tool in structural biology~\citep{cryo-revolution}, enabling three-dimensional reconstruction of macromolecular assemblies, often at near-atomic resolution. The Electron Microscopy Data Bank (EMDB~\citep{emdb}) now contains more than 56,000 density maps, providing a large resource for studying macromolecular structure and conformational variability. However, reconstructed cryoEM maps are three-dimensional scalar fields that do not by themselves specify atomic models, residue identities, secondary structure, or ligand-binding sites. Extracting these annotations typically requires substantial computational modeling and expert validation, and remains a bottleneck in converting maps into biological insight.

\begin{figure}[!ht]
    \centering
    \includegraphics[width=\textwidth]{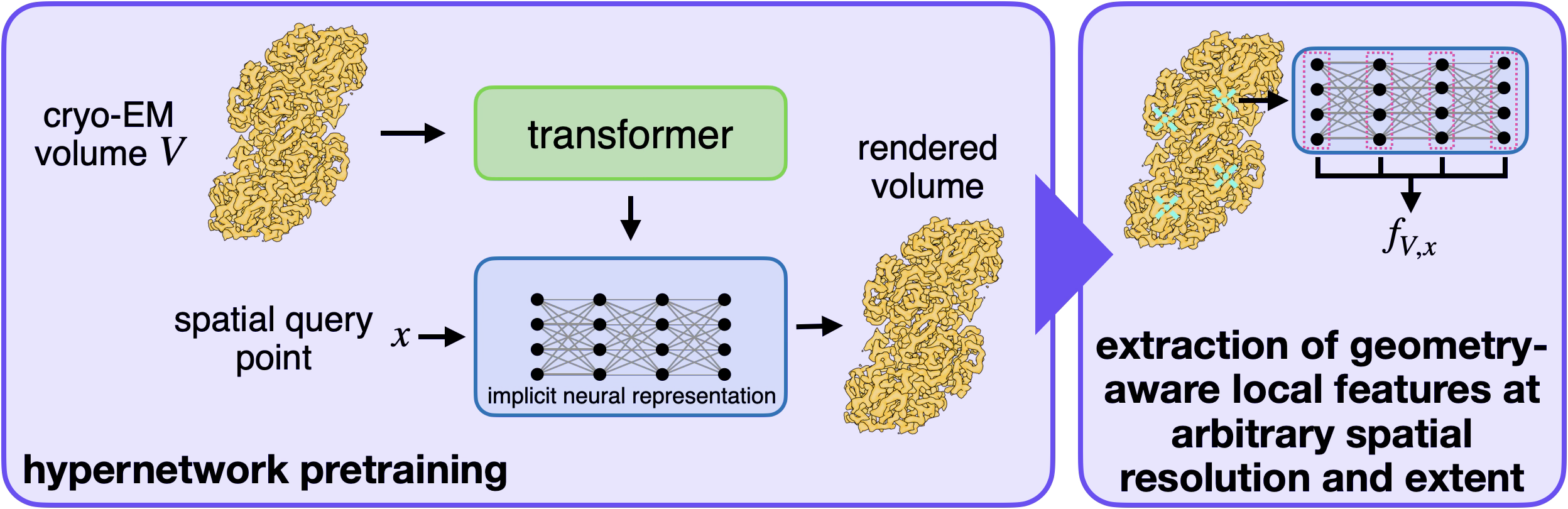}
    \caption{\textbf{Overview.} \modelname{} is a self-supervised framework for learning geometry-aware implicit neural representations of cryoEM density maps via hypernetworks. The generated representations expose local features at arbitrary spatial resolution, supporting downstream tasks on unseen protein structures.}
    \label{fig:teaser}
\end{figure}

CryoEM map interpretation relies on features spanning multiple spatial scales. Local density patterns support annotations such as secondary structure, nucleotide or protein-region identification, and residue-level model building, while larger-scale context is needed to interpret chain connectivity, molecular surfaces, domain organization, and interfaces between subunits~\citep{giri2024cryo2structdata, jamali2022graph}. Because the downstream-relevant features are not known in advance, a useful representation should expose information across scales, combining localized geometric detail with broader volumetric context.

Current deep learning methods for cryoEM map annotation, most notably 3D U-Nets, capture multi-scale spatial context and have been used for secondary-structure and nucleotide annotation~\citep{haruspex, emnuss}. However, these architectures operate on fixed voxel grids. Their feature fields are tied to the input discretization, and predictions at off-grid coordinates require interpolation rather than direct coordinate-conditioned evaluation.

Implicit neural representations (INRs) provide a scale-agnostic alternative to voxel-bound representations. They model a map as a coordinate-conditioned function $\mathcal{M}_\theta : [0,1]^3 \to \mathbb{R}$ that can be evaluated at arbitrary spatial coordinates~\citep{siren, nerf}, rather than only on a fixed lattice. This makes them well suited to cryoEM maps, which are discretized samples of an underlying continuous signal. Their main limitation is scalability: fitting one INR per map requires hundreds to thousands of optimization steps, and independently fitted INRs do not produce aligned intermediate features across maps. This prevents their direct use as shared feature extractors for downstream annotation. Hypernetworks that predict INR parameters from input data~\citep{ha2017hypernetworks,metasdf,functa,transinr} amortize this fitting cost, but cryoEM applications have focused primarily on reconstruction from particle images~\citep{cryohype,cryodrgn}, rather than transferable feature learning from reconstructed maps.

We introduce \modelname{} (Figure~\ref{fig:teaser}), a self-supervised framework that learns to map cryoEM volumes to INRs via a hypernetwork architecture. A transformer hypernetwork tokenizes a 3D density map and outputs the parameters of a compact INR in a single forward pass; this framework amortizes INR fitting across thousands of structures. By amortizing INR fitting into a feed-forward prediction, the hypernetwork produces instance-specific neural fields with implicitly aligned representations. This allows us to extract localized, coordinate-conditioned features that reside in a shared latent space across volumes. Crucially, \modelname{} is designed to \emph{augment} rather than replace existing voxel grid architectures. Because the generated INR is queryable at any continuous coordinate, its intermediate activations yield feature vectors that we use as auxiliary channels to a strong 3D nested U-Net baseline. This decomposition lets practitioners keep the multi-scale strengths of established discrete architectures while injecting a continuous, geometry-aware feature signal.

To validate the framework, we pretrain on $5439$ EMDB maps and evaluate the augmented U-Net on eight voxel-level annotation tasks, using a capacity-matched transformer-only autoencoder as a controlled baseline. The ablation lets us decompose the source of downstream gains into two components: the contribution of self-supervised transformer pretraining over a no-pretraining baseline, and the additional contribution of the INR formulation over the transformer alone.

In summary, our main contributions are:
\begin{itemize}
    \item \textbf{Amortized continuous representations for cryoEM at scale.}
    We introduce \modelname{}, a transformer-hypernetwork framework that generates per-volume INRs in a single forward pass. On $689$ held-out EMDB volumes, \modelname{} achieves an average normalized area under the Fourier shell correlation curve of $0.634 \pm 0.093$.
    \item \textbf{Coordinate-queryable features for downstream annotation.} We show that the intermediate activations of the generated INR can be queried at arbitrary spatial coordinates and used as auxiliary inputs to a 3D U-Net annotation head, improving voxel-level annotation accuracy on eight out of eight tasks against a volume-only baseline.
    \item \textbf{Decomposing the source of the improvement.} A capacity-matched transformer-only autoencoder ablation shows that self-supervised transformer pretraining accounts for the majority of the downstream gain, while the INR formulation typically contributes a smaller additional improvement and uniquely supports coordinate-conditioned, resolution-agnostic inference that the transformer alone cannot perform.
\end{itemize}

\section{Background and Related Work}
\label{sec:background}

\paragraph{Grid-Based Representation Learning in CryoEM.}
Cryo-electron microscopy determines macromolecular structure by reconstructing 3D maps from images of vitrified specimens acquired with an electron beam~\citep{nogales-review}. For favorable specimens, recent advances routinely produce near-atomic-resolution maps, including maps at 3\,\AA{} or better~\citep{lawsoncryoem}. Many biologically important maps, however, remain at intermediate or low resolution, especially in heterogeneous assemblies and in situ cryo-electron tomography, where cellular context is preserved at the cost of lower signal-to-noise ratio and resolution~\citep{berger2023cryo}. Map interpretation therefore spans a broad resolution range: high-resolution maps can support automated atomic model building with tools such as ModelAngelo~\citep{jamali2024automated}, while lower-resolution maps often require annotation, segmentation, fitting, or validation using a combination of automated methods and interactive tools such as Coot~\citep{coot} and ChimeraX~\citep{chimerax-structure-building}. Automated annotation methods, including secondary-structure prediction~\citep{haruspex, emnuss} and backbone tracing~\citep{deeptracer}, have historically relied on 3D U-Nets. While these methods automate important parts of map interpretation, they operate on discretized voxel grids. Consequently, their representations are tied to the input grid: increasing spatial resolution increases memory and compute cost, and predictions away from grid points require interpolation rather than direct coordinate-conditioned evaluation.

\paragraph{Implicit Neural Representations.}
Implicit neural representations (INRs) are coordinate-conditioned neural networks that model signals as continuous functions over space, and have been successfully applied to various domains including 2D images~\citep{inr-gan, liif}, 3D scenes~\citep{nerf, srn}, and audio~\citep{siren}. In practice, INRs are typically parameterized as lightweight multilayer perceptrons (MLPs), often augmented with sinusoidal activations~\citep{siren} or Fourier feature embeddings~\citep{fourier-features}.

The continuous formulation of INRs makes them particularly well-suited for cryoEM density maps, where biologically meaningful signals depend on fine-grained sub-voxel geometry. Prior work has explored INRs as a computational primitive for cryoEM, with~\citep{ranno-si} demonstrating their effectiveness for representing electron density. CryoDRGN~\citep{cryodrgn} and its extension cryoDRGN-AI~\citep{cryodrgnai} learn latent-conditioned INRs from heterogeneous particle images, while cryoAI~\citep{cryoai} fits an INR to reproduce observed projections under varying poses. However, these approaches are designed for recovering cryoEM densities from raw noisy images rather than learning transferable representations from reconstructed volumes.

Beyond reconstruction, INRs enable coordinate-conditioned feature extraction: intermediate activations can be queried at arbitrary spatial locations to produce feature vectors encoding local geometry at sub-voxel resolution. This perspective has been explored in recent work on neural fields as continuous feature volumes~\citep{hydif, kobayashi2022decomposing}. However, standard INRs are trained independently per datum, making them computationally expensive and resulting in representations that are not aligned across instances, limiting their use as scalable feature extractors.

\paragraph{Hypernetworks and Amortized INRs.}
To circumvent per-instance INR fitting, hypernetworks generate the parameters of a target network conditioned on an input~\citep{ha2017hypernetworks, babu2023hyperfields, transinr}, enabling amortized inference where a single forward pass produces instance-specific weights. In the INR setting, prior work such as MetaSDF~\citep{metasdf}, Trans-INR~\citep{transinr}, and HyperSound~\citep{hypersound} has shown that hypernetworks can efficiently generate high-fidelity implicit neural representations across diverse domains, including shapes, images, and audio. More recent work demonstrates their applicability to structured scientific data~\citep{hydif, babu2025acquiring}, while CryoHype~\citep{cryohype} applies hypernetworks to model heterogeneity in cryoEM reconstruction.

Hypernetworks induce a shared representation across instances by mapping inputs to a common function space. As a result, hypernetwork-generated INRs yield features that are implicitly aligned across volumes through the shared hypernetwork mapping, enabling localized, coordinate-conditioned features that reside in a shared latent space. This property makes neural fields a practical and scalable foundation for geometry-aware feature extraction across large collections of cryoEM volumes.

\paragraph{Intermediate Activations as Features.}
Extracting feature representations from intermediate network activations has a long history in computer vision, underpinning transfer learning in pretrained CNNs~\citep{decaf} and modern self-supervised learning~\citep{simclr, mae}. Recent work has observed that intermediate layers of INRs with sinusoidal activations encode localized spatial frequencies~\citep{siren} and geometric structure beyond what is required for pure reconstruction~\citep{functa}. A distinctive property of INR activations is \emph{spatial localization}, where querying the network at coordinate $x$ produces features intrinsically tied to the local geometry around $x$, in contrast with voxel encoders which require pooling for localized predictions. \modelname{} harnesses this property at scale, using coordinate-aligned INR activations from an amortized hypernetwork as features for voxel-level annotation without task-specific supervision during pretraining.

\section{Method}
\label{sec:method}

\begin{figure}
    \centering
    \includegraphics[width=\textwidth]{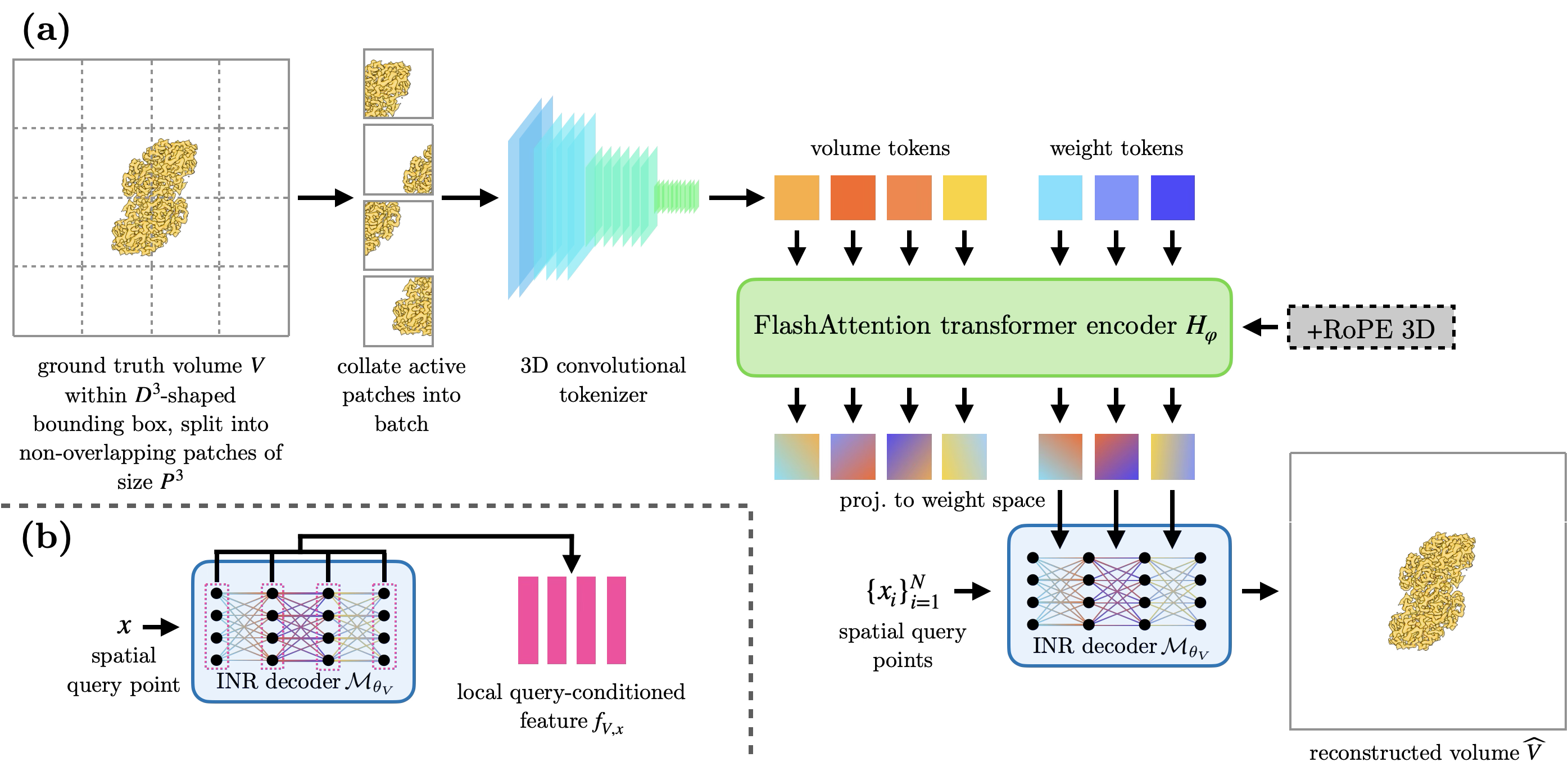}
    \caption{\textbf{\modelname{} architecture.} \textbf{(a)} A cryoEM volume $V$ occupies a $(D, D, D)$-shaped bounding box, split into non-overlapping patches of size $(P, P, P)$. Since the $(D, D, D)$ volume is naturally sparse with most voxels being background (set to a value of $0$), we collect the active patches containing non-background voxels into a batch and discard the rest. These patches are fed through a convolutional neural network that transforms each patch into a volume token. The volume tokens are passed into a transformer along with learnable weight tokens; 3D rotary positional embeddings inform the volume tokens of the patch that they came from. The contextualized weight tokens output by the transformer are then projected to the input volume-conditioned weight matrices $\theta_V$ of an INR decoder $\calM$, which can be queried at arbitrary spatial points in $[0, 1]^3$ to produce a reconstructed volume $\widehat{V}$. Density map of EMDB-0406~\citep{emdb0406} visualized with ChimeraX. \textbf{(b)} The INR naturally exposes an interface for fine-grained, coordinate-conditioned feature extraction. At inference, the transformer $\mathcal{H}_\phi$ produces weight matrices $\theta_V$ that parameterize the INR $\mathcal{M}_{\theta_V}$. For any query coordinate $x \in [0, 1]^3$, we extract the post-activation hidden states from each of the $K$ INR layers (pink) and stack them into a query-conditioned feature $f_{V, x} \in \mathbb{R}^{H \times K}$. Because $x$ is continuous, features can be queried at sub-voxel resolution and are spatially localized to the geometry around $x$ without interpolation.}
    \label{fig:architecture}
\end{figure}

\subsection{Data preprocessing}
\label{sec:data-preprocessing}
CryoEM volumes are scalar-valued voxel grids, where the value at each voxel represents electron density. Measurements are typically noisy, so we first denoise volumes via thresholding and removal of small connected components, then normalizing all voxel values to $[0, 1]$; this sets all background voxels to 0. We then pad the denoised volumes to a fixed $(D, D, D)$ size, where $D = 96$; any maps whose foreground voxels exceed this size are discarded. All maps are resampled to have a uniform resolution of 3\,\AA{} per voxel using the \texttt{resample} command in ChimeraX~\citep{chimerax}. We discuss our data curation and processing pipeline in further detail in \S\ref{sec:data-preparation}.

\subsection{Self-supervised INR pretraining}
\label{sec:method-pretraining}

While a cryoEM volume $V \in \mathbb{R}^{D \times D \times D}$ is natively a discrete tensor, we view it as a continuous function $V:[0, 1]^3 \to \mathbb{R}^+$, establishing a natural correspondence between the integer lattice $[D]^3$ and the continuous coordinate space $[0, 1]^3$. The goal of our self-supervised pretraining is to learn a coordinate-based INR $\mathcal{M}_{\theta_V}:[0,1]^3 \to \mathbb{R}^+$ such that $\mathcal{M}_{\theta_V} \approx V$ pointwise for each volume $V$.

A na\"ive approach would be to separately train an INR for each volume $V$. However, this precludes any generalization to new volumes and does not allow for synthesizing information across a diversity of protein structures. \modelname{} bypasses this by learning a global hypernetwork $\mathcal{H}_\phi$, parameterized by $\phi$. In other words, we learn a mapping $\mathcal{H}_\phi : V \mapsto \theta_V$ that observes a discrete cryoEM volume $V$ and directly predicts the parameters $\theta_V$ that render $V$ via the INR $\mathcal{M}_{\theta_V}$.\footnote{An \emph{atelier} is a workshop where a master artist trains many apprentices to produce work under the master's name. Just as the master produces trainees that render artworks, so too does the hypernetwork produce INRs that render cryoEM electron densities.}

The hypernetwork-to-INR formulation has been explored in prior work on 2D images and 3D shapes~\citep{transinr, metasdf, hypersound}. We adapt this paradigm to volumetric cryoEM data, with a training objective designed to recover the fine structural detail relevant to protein geometry. We sample the full dense grid of $N = D^3$ query points $\Omega \defeq \left\{i/(D-1) : i \in 0,\dots,D-1\right\}^3\subset[0, 1]^3$ and evaluate the INR over the grid to get a predicted volume $\widehat{V} = \calM_{\theta_V}(\Omega)$. In order to encourage the model to learn fine details associated with protein structures, our loss is the sum of the mean squared error loss between the volumes and the mean squared error of the 3D discrete orthonormal Fourier transform $\calF$ of the volumes, weighted by the square of the normalized frequencies $\xi\in[-1/2, 1/2)^3\eqdef\calF(\Omega)$ in order to upweight high-frequency components.

\begin{equation}
    \label{eqn:loss}
    \calL(\phi) = \bbE_{V\sim \calD}\left[\frac{1}{N}\sum_{x\in\Omega} |V(x) - \widehat{V}(x)|^2 + \frac{1}{N}\sum_{\xi\in\calF(\Omega)}|\xi|^2|\calF(V)(\xi) - \calF(\widehat{V})(\xi)|^2\right].
\end{equation}

Note that without the $|\xi|^2$ term, the Fourier loss would be equal to the spatial loss by Parseval's theorem. We use $\calD$ above to denote the data distribution.

\subsection{The \modelname{} architecture}
\label{sec:method-architecture}

The \modelname{} architecture, illustrated in Figure~\ref{fig:architecture}(a), consists of three main components: a tokenizer designed to handle high-dimensional sparse cryoEM volumes, a transformer encoder that generates positionally-aware volume and weight tokens, and an INR decoder that uses the transformed weight tokens to render volumes. Here we briefly describe each component, with further details in \S\ref{sec:architecture-details}.

\paragraph{The convolutional tokenizer.} Since we pad all volumes to the same $(D, D, D)$ shape and cryoEM volumes vary widely in size, most volumes will be sparsely populated with nonzero foreground voxels. To exploit this sparsity, we split the input volume $V$ into non-overlapping patches of size $(P, P, P)$ with $P = 6$, yielding a sequence of $(D/P)^3$ patches. The patches containing only background voxels are discarded, and the remaining active patches containing foreground voxels are processed in a batch by a 3D convolutional neural network into a sequence of $S$ latent volume tokens. 

\paragraph{The hypernetwork $\pmb{\mathcal{H}_\phi}$.} Learnable weight tokens are fed into a FlashAttention-enabled transformer~\citep{flash-attn, attention-is-all-you-need} along with the volume tokens output by the tokenizer. To retain global spatial context, we apply GPT-NeoX-style 3D rotary position embeddings (RoPE)~\citep{rope, gpt-neox-library} to the volume tokens in the transformer. The final sequence of weight tokens output by the transformer are imbued with spatial context from the volume tokens; we map the transformed weight tokens to weight matrices for an INR via a learnable linear projection.

\paragraph{The INR $\pmb{\mathcal{M}_\theta}$.} The INR takes the form of an MLP with ReLU activations and $K$ hidden layers of uniform width $H$. The weights are generated by the hypernetwork, while the biases are internal and not data-dependent. The INR is tasked with predicting the electron density at any point $x\in[0, 1]^3$ for the given volume $V$. To enable learning high-frequency functions, the spatial coordinate $x$ is first mapped into a higher dimensional space with a fixed Fourier encoding~\citep{nerf,fourier-features}.

\subsection{Local Feature Extraction}
\label{sec:method-features}

Typical self-supervised tasks map input sequences to tokens that are used directly to reconstruct a corrupted sequence, fundamentally limiting downstream reasoning to the resolution of the tokens. However, since our reconstruction is done with an INR $\cal{M}_\theta$ that can be queried at any continuous point in $[0, 1]^3$, we naturally are able to obtain latent representations of a volume $V$ at arbitrary resolution. Since $\mathcal{H}_\phi$ is trained to produce highly accurate structural reconstructions, the intermediate activations of $\mathcal{M}_{\theta_V}$ must encode rich, geometry-aware information about the local density neighborhood around any query point $x$, as illustrated in Figure~\ref{fig:architecture}(b). Given any query point $x$, we can obtain a query-conditioned feature $f_{V, x}\in\bbR^{H\times K}$ by extracting the post-activation hidden states from the INR evaluated at coordinate $x$. We can also obtain a feature for an arbitrary set of points $X = \{x_i\}_{i=1}^N$ by concatenating the activations at each point to get a region-conditioned feature $f_{V,X}\in\bbR^{N\times H\times K}$. 

This highly flexible extraction mechanism yields several critical advantages for structural biology. First, the feature $f_{V,x}$ is resolution-independent, meaning it can be queried at sub-voxel resolution regardless of the voxel spacing of the original map $V$. Second, it is inherently geometrically localized, as the spatial coordinates directly condition the activation cascade without requiring interpolation heuristics.

\section{Experiments}
\label{sec:experiments}

\subsection{Pretraining}
\label{sec:experiments-pretraining}
We pretrain \modelname{} on a library of 5439 cryoEM volumes, with validation and test sets of size 684 and 689 respectively, following the method described in \S\ref{sec:method-pretraining}. Our volumes are derived from the Cryo2StructData dataset~\citep{cryo2structdata}, which consists only of cryoEM volumes with resolved atomic structure (i.e., with corresponding fitted PDB maps). Following~\citep{cryobench,cryohype}, we evaluate the quality of our reconstruction by computing the area under the Fourier shell correlation curve (AUFSC) between ground truth $V$ and prediction $\widehat{V}$. We also report the \textit{normalized} AUFSC, which is always between 0 and 1 (1.0 indicating perfect reconstruction) and defined by us as
\begin{equation}
    \label{eqn:nAUFSC}
    \nAUFSC(V,\widehat{V})\defeq \frac{\AUFSC(V, \widehat{V})}{\AUFSC(V, V)}.
\end{equation}

\begin{figure}
    \centering
    \includegraphics[width=0.95\textwidth]{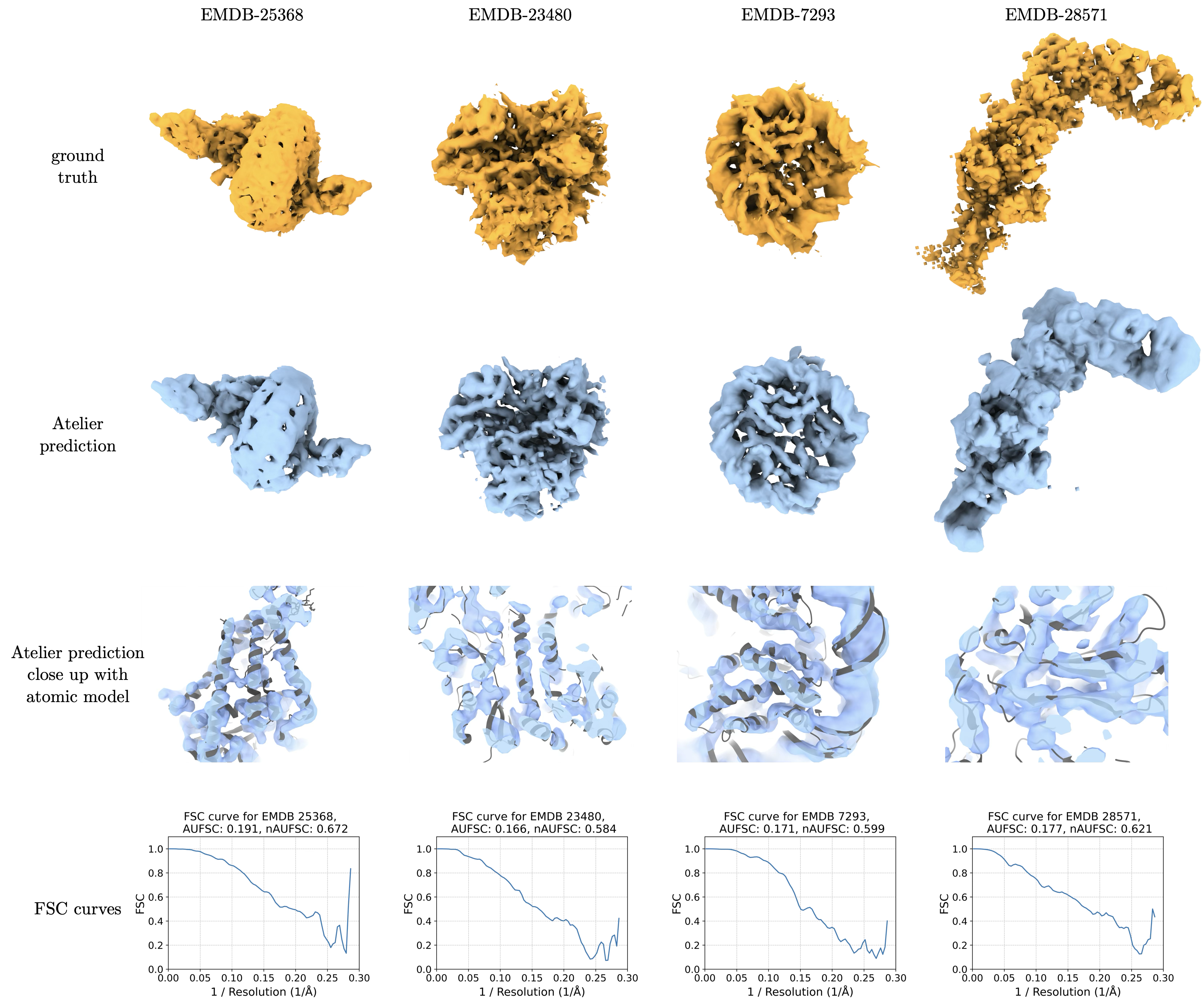}
    \caption{\textbf{\modelname{} successfully generalizes to unseen volumes.} Here we show ground truth (orange) and INR-rendered (blue) volumes for four representative samples in the test set, as well as the Fourier shell correlation curves between them. We also show close ups for the predicted densities overlaid over ground truth atomic models to demonstrate that \modelname{} is able to reconstruct near atomic-level detail. Note the biological diversity of structures; from left to right, the structures are a membrane-bound enzyme (EMDB-25368~\citep{emdb25368}), a nanobody bound to an HIV envelope glycoprotein (EMDB-23480~\citep{emdb23480}), a nucleosome--kinetochore complex (EMDB-7293~\citep{emdb7293}), and a fully assembled TCR/CD3/pMHC immune-recognition complex (EMDB-28571~\citep{emdb28571}).}
    \label{fig:sample-reconstructions}
\end{figure}

On 689 test volumes, we are able to obtain a mean$\pm$std nAUFSC of $0.634\pm 0.093$ (AUFSC: $0.180\pm 0.026$). We show qualitative and quantitative results of reconstructed test volumes in Figure~\ref{fig:sample-reconstructions}. In particular, as one of the examples, we show \modelname{}'s rendering of a fully assembled T-cell receptor (TCR), CD3, and peptide-MHC complex (EMDB-28571~\citep{emdb28571}). Many of the most biologically important macromolecules are large, multi-chain assemblies whose function depends on the relative arrangement of distinct subunits, often spanning both soluble and membrane-embedded regions. Faithfully reconstructing such complexes is a demanding test for a volumetric model, since the network must simultaneously recover distant subunits, asymmetric chain organization, and the contrast change between protein and detergent or lipid density. For EMDB-28571, \modelname{} recovers both the membrane-distal TCR$\alpha\beta$/pMHC recognition interface and the membrane-proximal CD3 signaling subunits in a single forward pass, despite the complex's asymmetric multi-chain architecture and embedded transmembrane region. The achieved nAUFSC is 0.621, comparable to the test set mean, demonstrating that reconstruction fidelity holds on large, heterogenous assemblies.

\begin{figure}
    \centering
    \includegraphics[width=\textwidth]{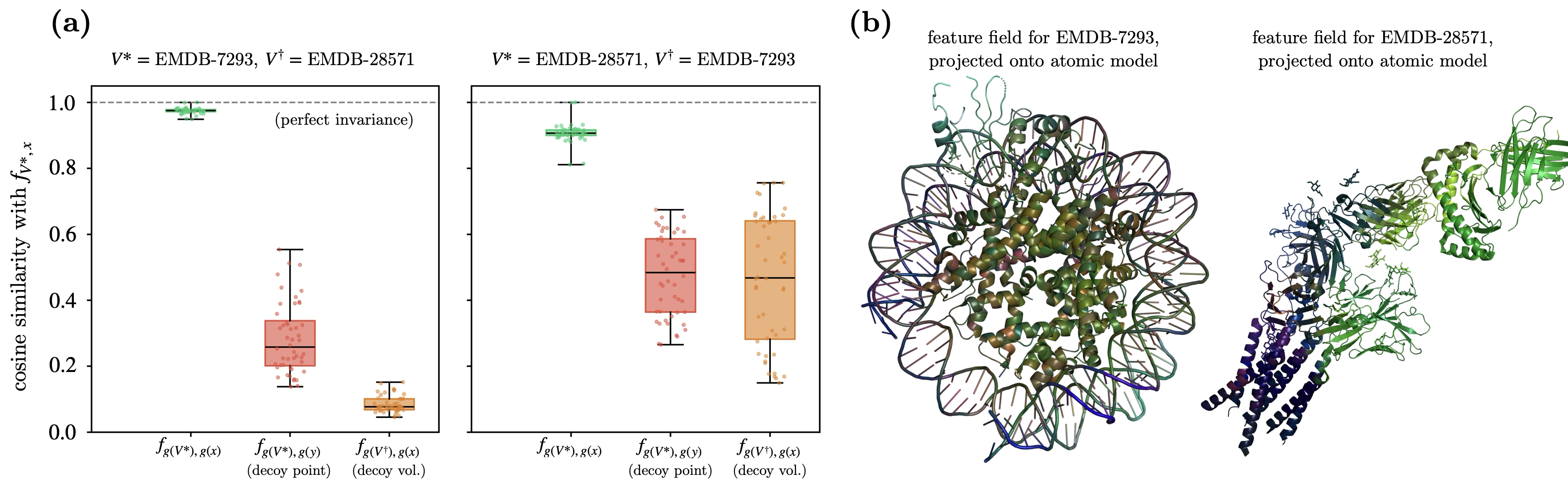}
    \caption{\textbf{\modelname{} generates meaningful local continuous features.} \textbf{(a)} After pretraining, \modelname{} is able to generate local features invariant to global symmetries due to its geometrically-driven pretraining task in two structures from Figure~\ref{fig:sample-reconstructions}. Box denotes minimum, quartiles, and maximum. \textbf{(b)} PyMol visualization of the same two structures evaluated in part (a), with atoms colored by trilinearly interpolating voxel features projected and scaled to $[0, 1]^3$ via PCA. Observe in the nucleosome (EMDB-7293) that the histones (center, green) primarily occupy a different color space than the surrounding DNA (purple) and centromere protein (upper left, teal). In the TCR/CD3/pMHC complex (EMDB-28571), there is considerably less variation within chains than across chains. Note, for example, the consistent coloring of the transmembrane region (bottom, purple).}
    \label{fig:meaningful-features}
\end{figure}

We perform two experiments to demonstrate that our pretraining results in geometrically meaningful local features via the extraction method described in \S\ref{sec:method-features}. During training, we augment all volumes by a random element in $g\in O_h$, the group of all 48 symmetries of the cube (generated by $90^\circ$ axis-aligned rotations and reflections). To demonstrate invariance of our local features, we sample a volume $V^*$ and a foreground point $x\in[0, 1]$, which we use to compute a localized feature $f_{V^*,x}$ from the pretrained model. We then compute the cosine similarity between $f_{V^*, x}$ and $f_{g(V^*), g(x)}$ for all $g\in O_h$; we expect the features to be aligned if the model has learned to generate local representations that are invariant to global symmetries. As negative controls, we also compute, for all $g\in O_h$, the cosine similarities between $f_{V^*, x}$ and $f_{g(V^*), g(y)}$ for a decoy foreground point $y\neq x$, as well as between $f_{V^*, x}$ and $f_{g(V^\dagger), g(x)}$ for decoy test volume $V^\dagger\neq V^*$. We visualize our results in Figure~\ref{fig:meaningful-features}(a) and observe that our model has learned approximately invariant local representations; we argue that this is a natural consequence of the pretraining task, which forces the model to learn geometrically informed features that are locally meaningful.

We also visualize the actual activations for two test proteins in Figure~\ref{fig:meaningful-features}(b) by computing features at all foreground voxels, projecting down to three dimensions and normalizing to $[0, 1]^3$ (which we interpret as RGB space) via principal components analysis, coloring each atom in the corresponding atomic model by trilinear interpolation of the RGB values on the voxel grid, and visualizing in PyMol~\citep{pymol}. We observe that the feature field is, in many cases, approximately constant within domains while varying between them, suggesting that our pretrained fine features carry some biologically useful signal and can partially segment novel proteins.

\subsection{Demonstrating the Locality of \modelname{} Fine Features}

To demonstrate the locality of the features that \modelname{} is able to extract from cryo-EM volumes, we performed the following study on the effect of local versus distant corruptions to a protein on \modelname{} activations. We collected 16 held-out test structures and independently sampled 16 foreground voxel coordinates each; the foreground voxels within a single structure are at least 12 voxels (= 36\,\AA) apart. For each of the 16 volumes, we generate 16 different corruptions where we zero out all voxels within an $R=6$ voxel radius, centered at one of the foreground points $x$, and measure the effect of this deletion on the post-ReLU activations at each layer of the INR generated by the trained \modelname{} model. As a matched control, we repeat the deletion at a second foreground point $y$ in the same structure, chosen from the remaining 15 points as the one whose enclosed mass most closely matches that of the ball at $x$, subject to the two balls not overlapping; that the total mass in the ball around our chosen $y$ is always within 25\% of the mass within the ball centered at $x$. Our hypothesis is that deleting the far neighborhood centered at $y$ has much less of an effect than deleting the local neighborhood at $x$. Note that the points within a volume are spaced far enough so that deleting a neighborhood around $y\neq x$ will not delete the point $x$ itself.

Our measure of how much deletions change the activations is given by:
\begin{equation}
    E_\ell (x; y, R) = \frac{\|a_\ell(x, V_{R, y}) - a_\ell(x, V)\|}{\|a_\ell(x,\varnothing) - a_\ell(x, V)\|},
    \label{eqn:deletion-perturbation}
\end{equation}
where $a_\ell(x, V)$ is the layer-$\ell$ post-ReLU activation of an INR generated from a cryo-EM volume $V$, and $V_{R,y}$ is the volume $V$ with a ball of radius $R$ centered at $y$ zeroed out. We use $\varnothing$ to denote the all-zeros volume. The numerator is therefore the difference in the activations between the deleted and original volume, while the denominator normalizes so that deleting the entire volume would have score 1; a larger score means a stronger perturbation in the $x$-localized features from deleting point $y$.

For each volume and each hidden layer, we compute the ratio of the average (over all 16 points) \textit{near} perturbation $E_\ell(x; x, R)$ and the average \textit{far} perturbation $E_\ell(x; y, R)$. We report the median value over all 16 volumes at all layers in Table~\ref{tab:locality-results}. We also report the same ratio for an \textit{untrained} instantiation of our model. As seen in Table~\ref{tab:locality-results}, local perturbations have between an $18\times$ and $34\times$ larger effect than distant ones for the trained model. Furthermore, for the untrained instantiation of \modelname{}, local and distant perturbations have nearly the same effect magnitude, indicating that this locality is \textit{learned} rather than an inherent property of the architecture.

\begin{table}
\caption{Per-layer median near/far average corruption effect (see Equation (\ref{eqn:deletion-perturbation})) over 16 holdout volumes (95\% bootstrap confidence interval) for trained and untrained \modelname{} model. For the trained model, perturbations local to a point change intermediate activations significantly more than distant perturbations, while for the untrained model, local and distant perturbations have nearly identical effect. This indicates that \modelname{} activations capture local information, and that this locality is learned.}
\label{tab:locality-results}
\centering
\setlength{\tabcolsep}{2pt}
\begin{tabular}{ccc}
\toprule
\textbf{layer} & \textbf{trained near/far ratio} & \textbf{untrained near/far ratio}\\
\midrule
layer 1 & 18.6 [13.7, 29.9] & 1.01 [0.98, 1.02]\\
layer 2 & 25.1 [16.3, 38.3] & 0.99 [0.93, 1.03]\\
layer 3 & 34.4 [23.6, 39.9] & 0.99 [0.96, 1.03]\\
layer 4 & 29.3 [17.7, 38.4] & 1.00 [0.98, 1.03]\\
\bottomrule
\end{tabular}
\end{table}

\subsection{Using Local Features for Downstream Property Prediction}
\label{sec:experiments-property-prediction}
To demonstrate the usefulness of the local features described in \S\ref{sec:method-features}, we perform an experiment where we adapt EMNUSS~\citep{emnuss} to predict various local properties for cryoEM volumes and augment the input to EMNUSS with our query-conditioned features. EMNUSS is a 3D nested U-Net~\citep{unetpp} architecture for secondary structure prediction that takes a $(D, D, D)$-shaped 3D cryoEM volume as input and was originally designed to densely predict a $(D, D, D, 3)$-shaped prediction of whether each voxel containing a backbone atom is part of a coil, helix, or strand (voxels not containing backbone atoms do not contribute to the loss).

Since our data consists only of cryoEM volumes with resolved atomic structure, we are able to extract eight different types of voxel-localized labels for our volumes, four of which are classification tasks (secondary structure, amino acid, molecule type, molecule class) and four of which are regression tasks (hydrophobicity, absolute SASA, relative SASA, crystallographic B-factor); we describe these labels in more detail in \S\ref{sec:data-preparation-labels}. We adopt the exact same architecture as EMNUSS, only changing the number of output channels depending on the task. Classification tasks are trained with standard cross-entropy loss and regression tasks are trained with mean squared error.

For each voxel containing an alpha-carbon, nucleotide glycosidic carbon, or other heavy non-water heteroatom, we extract two types of embeddings. The first type of embedding we extract for voxels with defined annotations is a fine-grained localized embedding obtained by sampling all voxels contained in the sphere of radius 3 voxels centered at the annotated voxel, extracting their corresponding hidden activations as described in \S\ref{sec:method-features}, mean pooling over the $K$ intermediate activations, and performing a Gaussian-weighted mean pooling over the voxels in the sphere so that features corresponding to voxels further away from the center of the sphere contribute less to the final $H$-dimensional feature; we pool activations from the sphere instead of only the single feature at the voxel itself in order to incorporate some local context. As these features are extracted from an INR trained to render the volume at fine detail, we expect them to encode highly localized information useful for biological annotation. The features are extracted from a frozen \modelname{} after pretraining.

To demonstrate the importance of these fine-grained local features from the INR, we extract a second type of embedding; these are coarse, patch-level embeddings for each annotated voxel for comparison. To do this, we retrain an autoencoder to perform the self-supervised volume reconstruction task without the INR. This transformer-only autoencoder uses the same volume tokenizer, but now has a transformer that only takes in volume tokens (without any weight tokens); the transformed volume tokens are projected to directly reconstruct the input volume at the corresponding patch; see \S\ref{sec:architecture-details-transformer-only} for details. For each annotated voxel, we can associate with it the $H$-dimensional transformed volume token corresponding to the patch that the voxel lives in as a coarse representation at the patch level. Since all annotated voxels within a patch will have identical representations, this is a much less fine-grained representation. For fair comparison, the latent dimension $H$ of the transformer-only autoencoder is the same as the hidden layer size $H$ of the INR in \modelname{}, and once again our features are extracted from the frozen transformer-only autoencoder after pretraining.

For each task, we retrain EMNUSS from scratch with four different types of inputs.\footnote{Molecule type and molecule class are two evaluations of the same model; see \S\ref{sec:data-preparation-labels}.} As a baseline, we input only the $(D, D, D, 1)$ volume into EMNUSS. To illustrate the efficacy of our INR-derived local features, we also retrain with the input augmented to a $(D, D, D, 1+H)$ multi-channel volume, with $H$ representing either the \modelname{} INR-derived local feature or the transformer-only autoencoder's coarse feature. These augmentations are applied to voxels with annotations; for voxels without annotations, we set the extra $H$ channels to zero. Since these augmented features, appended mostly at alpha-carbons, may provide a signal as to the shape of the protein backbone, as an additional control we also run an experiment where we append \textit{random} features to each annotated voxel. The random features have the same shape as the fine features and are generated by computing the global per-channel mean/std of the re-normalized frozen fine features over all voxels in all volumes, and attaching a random vector to each annotated voxel following this distribution; this guarantees that the first two moments of the random features match those of the \modelname{} fine features. Note that other than the dimension of the input channel, we use the exact same training setup for all three types of inputs to ensure a fair comparison.

We use the same train/validation/test splits as in the pretraining task, and report test set balanced accuracy (arithmetic mean of per-class recall) or relative error for each of the three inputs in Table~\ref{tab:property-prediction-results}. We also report the Matthews Correlation Coefficient (MCC)~\citep{MCC}, another robust metric for imbalanced multiclass classification, in Table~\ref{tab:mcc-results} in the Appendix. We observe that augmenting with our local features from the INR pretraining outperforms the augmentation with coarse patch-level features in seven out of eight tasks, and outperforms the volume-only vanilla EMNUSS baseline in eight out of eight tasks. We attribute this to the fact that the pretraining task forces the learning of rich representations capturing local geometry, which improves downstream localized property prediction. Our results suggest that the INR activations from \modelname{} contain additional structural information orthogonal to that which is contained in the cryoEM density alone. We emphasize once more that we are augmenting with \emph{frozen} features after pretraining, indicating that purely geometric information is able to improve performance on biological annotation. Furthermore, our transformer-only autoencoder achieves an average per-volume nAUFSC of $0.870\pm 0.061$ (AUFSC: $0.248\pm0.017$); despite outperforming \modelname{} on reconstruction, it typically underperforms on property prediction, highlighting the importance of the fine features extracted by the INR in \modelname{}.

\begin{table}
\caption{Comparison on property prediction performance on the test set for (a) vanilla EMNUSS that takes the single-channel volume alone as input, (b) EMNUSS with inputs augmented with coarse features as additional channels, and (c) EMNUSS with inputs augmented with fine features as extra channels. Metrics are averaged over all labeled voxels in the test set; a more detailed, per-volume statistical analysis is reported in \S\ref{sec:emnuss-details}. Augmentation with fine features outperforms the volume-alone model in eight out of eight tasks, and outperforms the coarse model in seven out of eight tasks.}
\label{tab:property-prediction-results}
\centering
\setlength{\tabcolsep}{3pt}
\begin{tabularx}{\textwidth}{
    p{2.2cm}
    S[table-format=1.3]
    S[table-format=1.3]
    S[table-format=1.3]
    S[table-format=1.3]
    S[table-format=1.3]
    S[table-format=1.3]
    S[table-format=1.3]
    S[table-format=1.3]
}
\toprule
& \multicolumn{4}{c}{\textbf{classification (balanced accuracy $\uparrow$)}} 
& \multicolumn{4}{c}{\textbf{regression (relative error $\downarrow$)}} \\
\cmidrule(lr){2-5} \cmidrule(lr){6-9}
network input & {sec.\ struc.} & {amino} & {mol.\ type} & {mol.\ class.} 
& {hydro.} & {abs.\ SASA} & {rel.\ SASA} & {B-fact.} \\
\midrule
volume alone 
& 0.729 & 0.207 & 0.232 & 0.903 & 0.806 & 0.475 & 0.425 & 0.369 \\
w/ coarse feats. & 0.788 & {\bfseries 0.249} & 0.296 & 0.974 & 0.761 & 0.397 & 0.357 & 0.381 \\
w/ rand. feats. & 0.755 & 0.231 & 0.283 & 0.968 & 0.762 & 0.392 & 0.338 & 0.399 \\
w/ fine feats. & {\bfseries 0.797} & 0.235 & {\bfseries 0.300} & {\bfseries 0.976} & {\bfseries 0.752}
 & {\bfseries 0.386} & {\bfseries 0.331} & {\bfseries 0.363} \\
\bottomrule
\end{tabularx}
\end{table}

\subsection{Case Study: Comparing \modelname{} Features Across Related Structures}
\begin{figure}
    \centering
    \includegraphics[width=\textwidth]{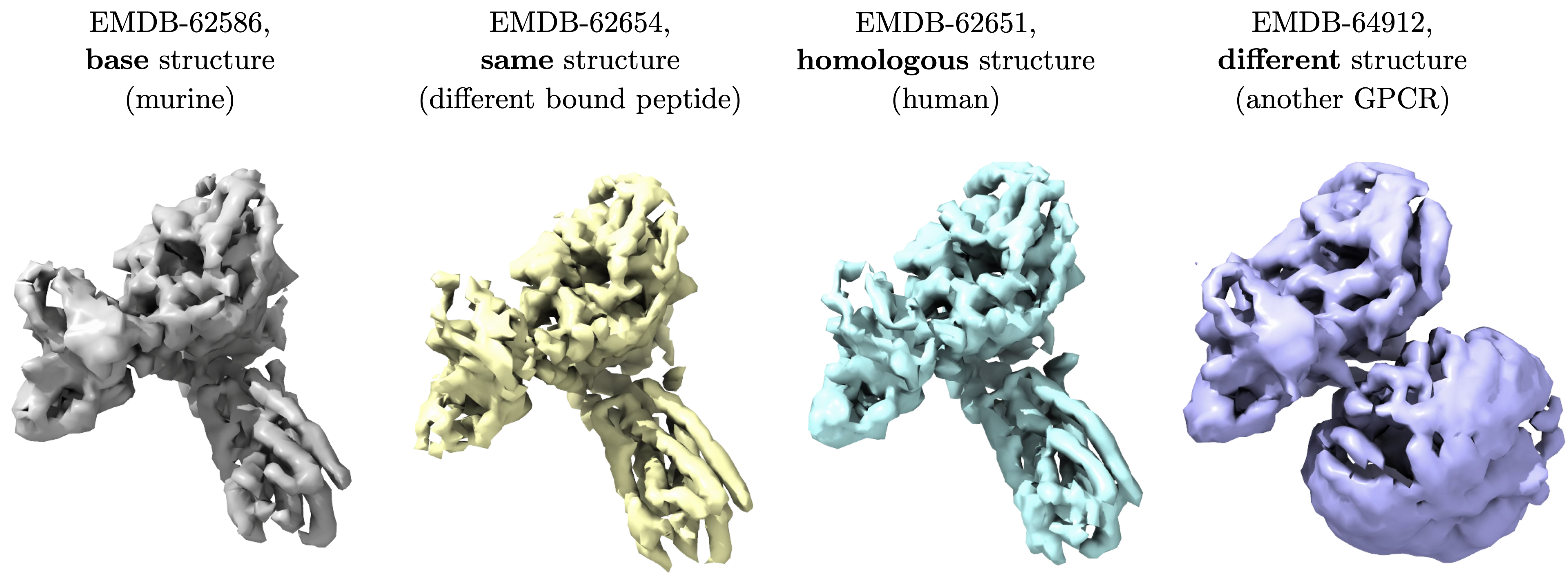}
    \caption{Illustration of the family of four complexes compared in Table \ref{tab:homology-results}.}
    \label{fig:gprotein-quartet}
\end{figure}

To investigate if the hypernetwork nature of \modelname{} is able to learn representations that are consistent across different volumes, we ran the following case study on the following quartet of four G-protein coupled receptors (GPCRs), all of which were not present in the training data, to examine if \modelname{} features are conserved across evolutionarily related structures:
\begin{enumerate}
    \item EMDB-62586~\citep{emdb62586}: we call this the \textbf{base} structure, a mouse TLQP21 bound to mouse C3aR in complex with Go;
    \item EMDB-62654~\citep{emdb62651_62654}: this is the exact same structure just with a different bound peptide, EP67 bound mouse C3aR in complex with Go; we call this the \textbf{same} structure (only the peptide differs);
    \item EMDB-62651~\citep{emdb62651_62654}: human structure with homologous protein sequences, structure of EP67 bound human C3aR in complex with Go; we call this the \textbf{homologous} structure;
    \item EMDB-64912~\citep{emdb64912}: a human Neurotensin Receptor 1 (hNTSR1)-Gi1 complex in nucleotide-free NC state 3; the NTSR1 and C3aR receptors are both G protein coupled receptors, but are more distantly related than human and mouse C3aR; we call this the \textbf{different} structure.
\end{enumerate}
We aligned all four complexes (pictured in Figure \ref{fig:gprotein-quartet}, resampled to 3~\AA{} voxels, and selected 80 coordinates that are foreground (nonzero) in all four structures simultaneously and mutually at least 4 voxels (12~\AA) apart. At each point we computed a localized feature exactly as in our EMNUSS experiments, using the same pretrained \modelname{} model, and then subtracted each volume's mean feature as a control for volume identity. In Table~\ref{tab:homology-results}, we report the cosine similarity at corresponding points between the base and same structures, base and homologous structures, and base and different structures. We observe that the cosine similarity decreases as structural relatedness increases, with the closely related structures (base, same, and homologous) clearly separated from the distantly related receptor (different).

\begin{table}
\caption{Cosine similarities between \modelname{} fine features at corresponding coordinates between four different structures (mean $\pm$ std. over 80 sampled points). The results suggest that at corresponding locations, the activations for more similar structures are more similar than those of distantly related structures, even after spatial alignment. As negative controls, we also computed the cosine similarities between the 80 points in the base structure and the same 80 points in a completely unrelated decoy volume (EMDB-28571, a TCR/CD3 complex, cosine similarities 0.450 $\pm$ 0.417, as well as computing the cosine similarity between a base point and one mismatched point each from the same/homologous/different structures (0.021 $\pm$ 0.441).}
\label{tab:homology-results}
\centering
\setlength{\tabcolsep}{2pt}
\begin{tabular}{cc}
\toprule
comparison & fine feature cosine similarity between volumes (mean $\pm$ std.)\\
\midrule
base vs. same & 0.949 $\pm$ 0.059\\
base vs. homolog & 0.926 $\pm$ 0.095\\ 
base vs. different & 0.774 $\pm$ 0.321\\
\bottomrule
\end{tabular}
\end{table}

\section{Conclusion}

By training a transformer hypernetwork to generate implicit neural representations of high-resolution cryoEM volumes, \modelname{} enables the extraction of geometrically-rich, localized feature vectors that can be queried at any spatial coordinate. We demonstrate the efficacy of these features by augmenting a baseline secondary structure prediction model with our coordinate-conditioned activations, showing consistent improvements on eight out of eight voxel-level property prediction tasks. We note several boundaries of our current evaluation that offer clear directions for future work in \S\ref{sec:limitations}.

Beyond the tasks studied here, \modelname{}'s localized features are naturally suited to annotating functional interfaces in large protein complexes. In assemblies like the immune synapse, distinguishing the precise contacts between TCR, peptide, and MHC subunits requires sub-voxel, side-chain-level precision. Discrete voxel grids risk losing fine geometric signal at this scale. \modelname{}'s continuous features provide a strong foundation for future work in interface prediction, binding-site identification, structure-guided protein engineering, and the next generation of learned representations of cryoEM density maps.

\bibliographystyle{unsrtnat}
\bibliography{refs}


\appendix

\section{Data Preparation}
\label{sec:data-preparation}
We base our data curation pipeline heavily on the method in Cryo2StructData~\citep{cryo2structdata}, starting with a list of 7392 curated EMDB IDs with corresponding fitted PDB maps. We downloaded all density maps from the EMDB~\citep{emdb} and corresponding \texttt{.pdb} files from the RCSB Protein Data Bank~\citep{pdb}. All density maps are resampled to a uniform voxel size of 3\,\AA{}$\times$3\,\AA{}$\times$3\,\AA{} using the \texttt{resample} command in ChimeraX.

\subsection{Denoising CryoEM Volumes}
\label{sec:data-preparation-denoi}

\begin{figure}
    \label{fig:denoising}
    \centering
    \includegraphics[width=\textwidth]{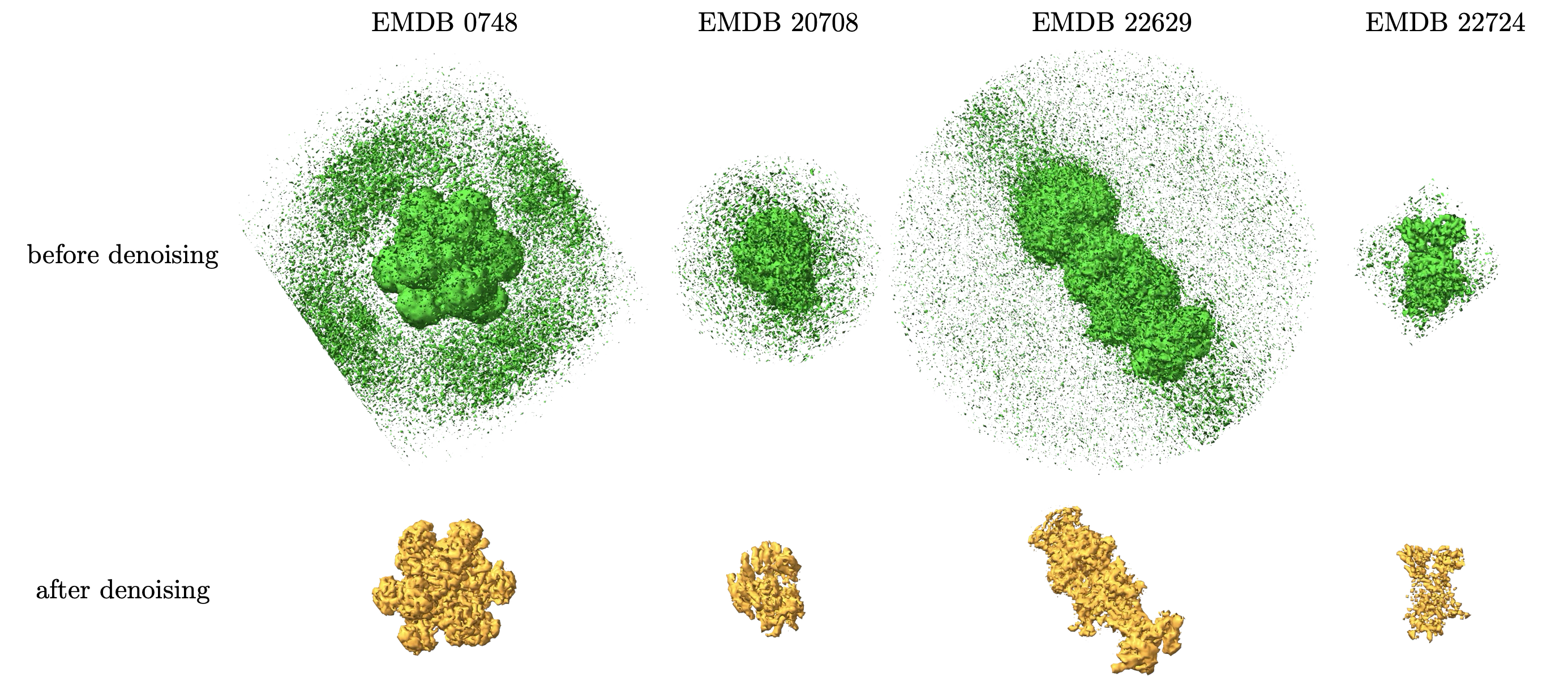}
    \caption{Illustration of the effect of our denoising method on four uncurated cryoEM volumes. By thresholding out the ``dust'' present in raw deposited maps, it becomes straightforward for the tokenizer component of \modelname{} to only tokenize patches containing real signal.}
\end{figure}

Reconstruction of 3D electron densities from raw cryoEM images is a notoriously difficult problem, with the raw measurements extremely low signal-to-noise ratio~\citep{singer}. As a result, even deposited maps can be highly noisy and contain large amounts of ``dust''. In order to allow our tokenizer to ignore background patches that contain only noise, we wish to threshold out all noise to a background value of zero. Various competing methods have been developed before, some deep-learning based~\citep{deepemhancer,emready,crefdenoiser} and some using classical signal processing techniques~\citep{lafter}. For simplicity, we devise our own denoising algorithm, which performs the following steps on the 3\,\AA{} resampled map:
\begin{enumerate}
    \item We first apply Otsu thresholding~\citep{otsu} to mask out the background, then compute the volume reduction of the tight bounding box around the foreground. If the reduction is less than 10\% (indicating an unusually noisy background), we instead sweep 50 threshold values between the minimum and the 50th percentile voxel value and select the one that maximizes the discontinuity in data reduction to threshold out the background.
    \item To further remove ``dust'', we compute the size of all connected components (using 26-connectivity) in the volume after thresholding and remove anything smaller in size than 1\% of the largest connected component.
\end{enumerate}
Since the voxel values in cryoEM maps are unitless, we linearly scale all voxel values to $[0, 1]$ after denoising. In Figure~\ref{fig:denoising}, we show four representative samples before and after applying our denoising method. Denoised volumes are saved to disk and padded to size $(96, 96, 96)$ by the dataloader during training and inference; any volumes exceeding this size are discarded, leaving a total of 5439/684/689 usable train/validation/test samples. Splits were generated randomly before filtering for size.

\subsection{Label Generation}
\label{sec:data-preparation-labels}

Since all cryoEM volume in our dataset have corresponding fitted PDB atomic structures, we are able to align atomic coordinates with the cryoEM voxel grid and label individual voxels with properties derived from the PDB files. For each cryoEM volume, we compute the following labels.
\begin{itemize}
    \item Secondary structure (classification, 3 labels): at voxels containing alpha carbons, we label the secondary structure corresponding that residue as determined by ChimeraX.
    \item Amino acid (classification, 20 labels): at voxels containing alpha carbons, we label the identity of the corresponding residue as read from the PDB file.
    \item Molecule type (classification, 22 labels): an extension of secondary structure; at voxels containing either alpha carbons, nucleotide glycosidic carbons, or other heavy non-water heteroatoms, we label them accordingly (20 labels for amino acids, 1 for any nucleotide, and 1 for any heteroatom).
    \item Molecule class (classification, 3 labels, same model as molecule type): We do not train a separate model for this task; instead, we remap the molecule type predictions by grouping all amino acid labels into a single label, so that we may evaluate the molecule type-trained model's ability to distinguish between protein, nucleotide, and other.
    \item Hydrophobicity (regression): at voxels containing alpha carbons, we label with the Kyte-Doolittle hydrophobicity~\citep{hydrophobicity} of the corresponding residue.
    \item Absolute SASA (regression): at voxels containing alpha carbons, we label with the absolute solvent accessible surface area of the corresponding residue as computed by FreeSASA~\citep{freesasa}.
    \item Relative SASA (regression): at voxels containing alpha carbons, we label with the absolute solvent accessible surface area of the corresponding residue as computed by FreeSASA, divided by the empirical maximum possible SASA for each residue type as given by~\citep{rel-sasa}.
    \item B-factor (regression): at voxels containing alpha carbons, we label with the mean crystallographic B-factor, as read from the PDB file.
\end{itemize}
For each label, all voxels that do not contain alpha-carbons (or nucleotides and heteroatoms, for molecule type) are assigned a background label.

\section{Architecture and Training Details}
\label{sec:architecture-details}
\subsection{\modelname{} architectural details}
\label{sec:architecture-details-hypernetwork}

The total number of trainable parameters of \modelname{} with the hyperparameters described below is 38,332,806. Our architecture is adapted in part from that of Trans-INR \cite{transinr}.

\paragraph{The convolutional tokenizer.} Our dataloader pads all volumes to a size of $(96, 96, 96)$, but most proteins do not take up this whole volume. To exploit this sparsity, we divide the voxel grid into patches of size $(6, 6, 6)$ for a total of 4096 patches. Patches which only contain background voxels are discarded, while the remaining active patches are collected into a single batch and processed by a shared 3D convolutional neural network that outputs a $d/2$-dimensional token representing each active patch, where $d=256$ is the latent embedding dimension of the transformer. Across the dataset, the median number of active patches is 236, meaning that 95\% of patches are discarded.

\paragraph{Magnitude Invariant Parametrization encoder.} Before entering the transformer, both the volume tokens and the weight tokens are passed through a Magnitude Invariant Parametrization encoder (MIP~\citep{mip}):
\begin{equation}
    \label{eqn:mip}
    \MIP:x\mapsto [\cos(\pi x/2), \sin(\pi x / 2)].
\end{equation}
We found in early experiments that this improved training convergence. Note that this doubles the dimension of each token before feeding into the transformer.

\paragraph{The hypernetwork \pmb{$H_\phi$}.} Our hypernetwork is a transformer encoder with latent dimension 256, 24 attention layers, 8 attention heads, a head dimension of 64, and a feedforward hidden dimension of 2048 with GeLU activations and a dropout rate of $0.1$. Rotary positional embeddings are applied independently per spatial axis using the integer patch grid coordinates $(i, j, k)$. Each head dimension is divided into three equal groups $(x, y, z)$. Each group uses 20 rotary dimensions (60 total out of 64; the remaining 4 are left unrotated), with base frequency 10,000.

\paragraph{INR weight generation.}For each of the $K+1$ INR layers, we maintain a set of learned \emph{weight tokens}: one $(d/2)$-dimensional token per output neuron in that layer, initialized from $\mathcal{N}(0,1)$. These tokens are data-independent and shared across all input volumes. At each forward pass they are concatenated with the volume tokens along the sequence dimension, jointly MIP-encoded (Equation~\ref{eqn:mip}), and processed by the transformer. Because the weight tokens carry no spatial meaning, they are assigned position $(0,0,0)$ for the purpose of computing rotary embeddings.

After the transformer, the output positions corresponding to the weight tokens are extracted. For INR layer $i$ with $n_i^{\mathrm{out}}$ output neurons and $n_i^{\mathrm{in}}$ input features, the $n_i^{\mathrm{out}}$ extracted $d$-dimensional tokens are passed through a dedicated learnable linear projector $\mathbf{P}_i \in \mathbb{R}^{n_i^{\mathrm{in}}\times d}$. The resulting rows are $\ell_2$-normalized and scaled by a learned per-layer scalar $\alpha_i$ (initialized to $0.1$) to decouple weight direction from magnitude:
\begin{equation}
    \label{eqn:weight-gen}
    W_i^{(j)} = \alpha_i \cdot \frac{\mathbf{P}_i \mathbf{h}_j^{(i)}}{\bigl\|\mathbf{P}_i \mathbf{h}_j^{(i)}\bigr\|_2},
\end{equation}
where $\mathbf{h}_j^{(i)} \in \mathbb{R}^d$ is the transformer output for the $j$-th weight token of layer $i$. The $n_i^{\mathrm{out}}$ rows $W_i^{(j)}$ are stacked to form the weight matrix for layer $i$.

Biases are not generated by the hypernetwork. Each INR layer instead has a learned bias vector shared across all inputs, initialized uniformly in $\bigl[-1/\sqrt{n_i^{\mathrm{in}}},\,1/\sqrt{n_i^{\mathrm{in}}}\bigr]$.

\paragraph{The INR decoder $\calM_\theta$} Instead of taking raw spatial coordinates as input, the INR first encodes spatial query points $x\in[0, 1]^3$ with a fixed Fourier feature encoding:
\begin{equation}
    \gamma: x \mapsto \bigl[\sin(2\pi f_k x_i),\, \cos(2\pi f_k x_i)\bigr]_{\substack{k=1,\dots,16 \\ i\,\in\,\{1,2,3\}}} \;\in\; \mathbb{R}^{96},
\end{equation}
where $\{f_k\}_{k=1}^{16}$ are log-spaced in $[1, 256]$. With each forward pass, the INR is evaluated at all $96^3$ grid points to render the full volume.

\subsection{\modelname{} pretraining details.} 
\label{sec:atelier-pretraining-hyperparameters}
\modelname{} was trained using AdamW~\citep{adamw} with a $(\beta_1,\beta_2)=(0.90, 0.95)$ and a weight decay of $0.01$. We trained for 5000 total epochs with a linear warmup of 10 epochs to a learning rate of 0.001 followed by a cosine decay to 0.0001. Training was distributed over 8 H100 GPUs with an effective batch size of 64 volumes per batch. Gradients were clipped at 1.0. During training, we augment the volumes by a random element of the octahedral group $O_h$. Total wall time for pretraining was approximately 90 hours.

\subsection{Transformer-only ablation}
\label{sec:architecture-details-transformer-only}
\begin{figure}
    \centering
    \includegraphics[width=\textwidth]{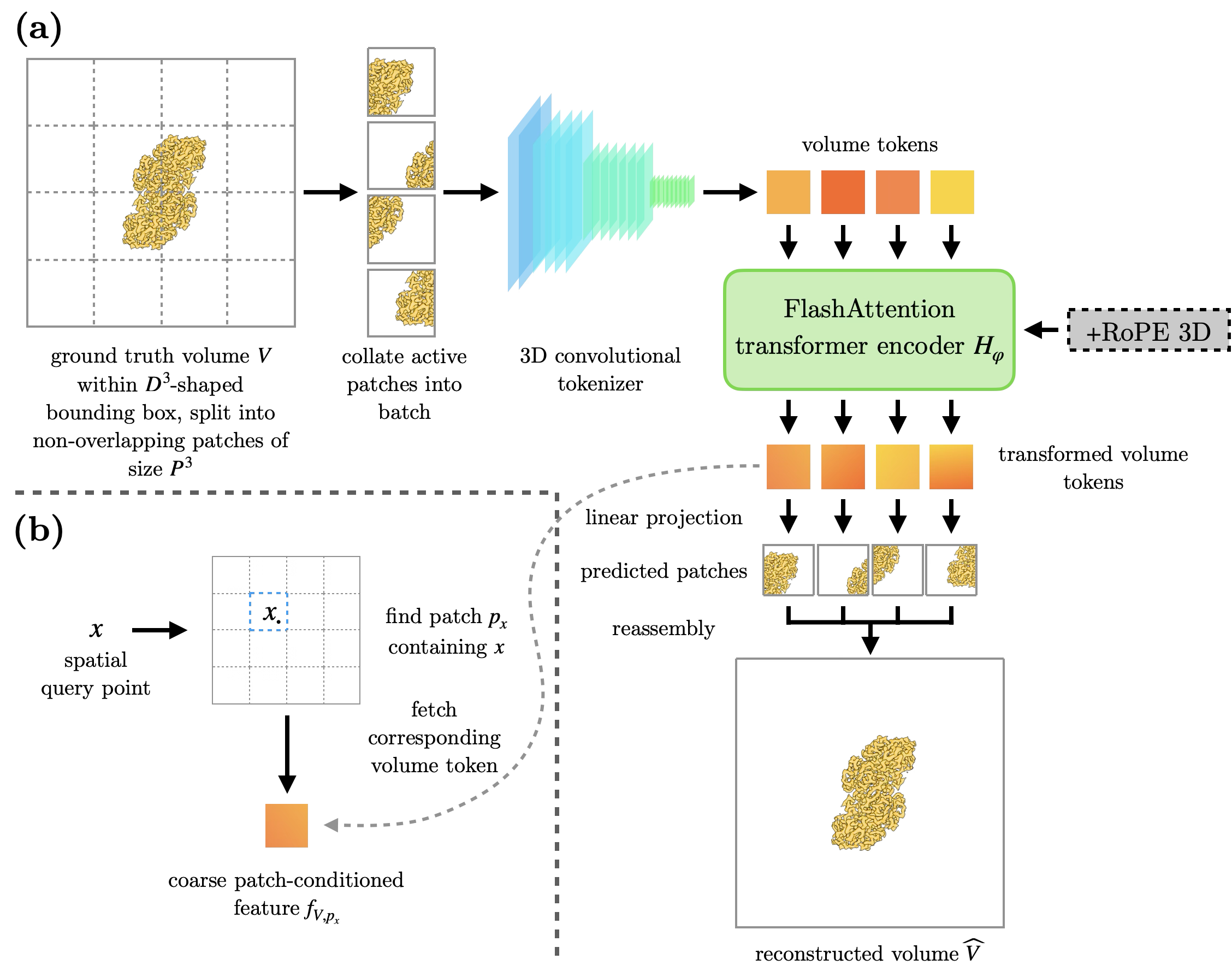}
    \caption{\textbf{The transformer-only ablation architecture.} \textbf{(a)} As an ablation, we train a transformer-only model that directly maps per-patch volume tokens to patch reconstructions through a learnable linear projection. \textbf{(b)} Coarse patch-level features can be extracted by returning the transformed volume token corresponding to the patch $p_x$ containing a query point $x$. Note that all query points within the same patch will have the same coarse feature.}
    \label{fig:transformer-only-ablation}
\end{figure}

To highlight the importance of the fine-grained features exposed by the hypernetwork structure of \modelname{}, we perform an ablation where we train a transformer-only variant of our architecture to reconstruct cryoEM densities, illustrated in Figure~\ref{fig:transformer-only-ablation}(a). The volume tokenizer and transformer remain exactly the same, except the transformer no longer accepts weight tokens. Instead, the transformed volume tokens are directly used to reconstruct the volume at the corresponding patch using a learned linear projection. The patches are then reassembled and the loss is computed as in Equation~\ref{eqn:loss}. We train this ablated model with exactly the same hyperparameters as \modelname{}.

\subsection{EMNUSS property prediction experiment details}
\label{sec:emnuss-details}

\paragraph{Architecture.}
We use EMNUSS~\citep{emnuss}, a 3D nested U-Net~\citep{unetpp} architecture for per-voxel property prediction. It takes a $C$-channel $(D, D, D, C)$ volume as input and outputs a $(D, D, D, k)$ volume for $k$-class classification or $(D, D, D, 1)$ for regression. For all three variants, we use the exact same implementation as the original EMNUSS codebase, only changing the input and output dimensions. 

\paragraph{Training.}

The training setup is identical for all three input variants (volume-only, augmentation with coarse features, augmentation with fine features) in the property prediction experiment from \S\ref{sec:experiments-property-prediction}. We use AdamW~\citep{adamw} with $(\beta_1, \beta_2) = (0.90, 0.999)$ and a weight decay of 0.01. We train for 300 total epochs with a linear warmup of 10 epochs to a learning rate of 0.001 followed by a cosine decay to 0.0001. Remaining training hyperparameters (weight decay, batch size, gradient clipping) are exactly the same as in \modelname{} pretraining as described in \S\ref{sec:atelier-pretraining-hyperparameters}, with the exception that we do not train with random rotations from $O_h$. Classification tasks are trained with cross-entropy loss with the background class excluded and evaluated by accuracy over non-background voxels. Regression tasks are trained with MSE computed only over non-background voxels and evaluated by the mean absolute relative error over non-background voxels, reported as ``relative error'' in Table~\ref{tab:property-prediction-results}. We use the same train/validation/test splits as \modelname{} pretraining.

\paragraph{Fine feature extraction (\modelname{}).}
For each annotated voxel at position $x$ we:
\begin{enumerate}
    \item collect all voxel positions within a sphere of radius 3 voxels centered at $x$;
    \item query the \emph{frozen} best-checkpoint \modelname{} INR at each sphere voxel and extract the $K{=}4$ post-ReLU hidden-layer activations, each $H{=}256$-dimensional;
    \item mean-pool across the $K$ layers to obtain one $H$-dimensional vector per sphere voxel;
    \item perform a Gaussian-weighted mean-pooling across sphere voxels with $\sigma = 1.5$ voxels, so voxels further from $x$ contribute less to the voxel-pooled representation.
\end{enumerate}
This yields one 256-dimensional feature per annotated voxel, precomputed offline from the frozen \modelname{} checkpoint.

\paragraph{Coarse feature extraction (transformer-only ablation).} For each annotated voxel at position $x$, retrieve the $H=256$-dimensional transformed volume token of the $6{\times}6{\times}6$ voxel patch $p_x$ containing $x$ from the frozen best-checkpoint transformer-only autoencoder (Figure~\ref{fig:transformer-only-ablation}(b)). Consequently, all voxels within the same patch share an identical representation.

\paragraph{MCC for Classification Tasks}
In Table~\ref{tab:mcc-results}, we report the Matthews correlation coefficient, a robust metric for highly imbalanced classification tasks~\citep{MCC, chicco-mcc} for classification experiments reported in Table~\ref{tab:property-prediction-results}. 
\begin{table}
\caption{Matthews correlation coefficient (MCC $\uparrow$) on the classification tasks from Table~\ref{tab:property-prediction-results}. Metrics are averaged over all labeled voxels in the test set.}
\label{tab:mcc-results}
\centering
\begin{tabular}{
    l
    S[table-format=1.3]
    S[table-format=1.3]
    S[table-format=1.3]
    S[table-format=1.3]
}
\toprule
network input & {sec.\ struc.} & {amino} & {mol.\ type} & {mol.\ class.} \\
\midrule
volume alone     & 0.607 & 0.186 & 0.223 & 0.756 \\
w/ coarse feats. & 0.681 & \bfseries{0.227} & 0.293 & 0.940 \\
w/ rand. feats   & 0.675 & 0.205 & 0.272 & 0.935 \\
w/ fine feats.   & \bfseries{0.705} & 0.217 & \bfseries{0.294} & \bfseries{0.950}\\
\bottomrule
\end{tabular}
\end{table}

\paragraph{Per-volume property prediction results.}
The property prediction metrics reported in Table~\ref{tab:property-prediction-results} are pooled over all labeled voxels in the test set. However, since voxels within a volume may be correlated, pooled results do not ensure consistency of improvements across volumes. We therefore run the following analysis:
\begin{enumerate}
    \item For each of the tasks, we compute the \textit{per-volume} metric with the fine features and the three baselines (volume-alone, coarse features, random features). We then compute the difference between the fine metric and three baselines for each volume; differences are oriented so that a positive value means the fine model is better.
    \item In Table~\ref{tab:per-volume-property-prediction-results}, for each comparison within each task, we report the Wilcoxon signed-rank test $p$-value, the Holm-Bonferonni corrected $p$-value to account for multiple testing, the win rate (fraction of maps on which the final map is better), and the Hodges-Lehmann effect size with a 95\% distribution-free confidence interval (positive means fine features are better). We see that for most comparisons, the \modelname{} fine features significantly outperform baselines.
\end{enumerate}
\begin{table}
\centering
\caption{Per-volume comparison on fine \modelname{} features against baselines. We see that for amino acid prediction, coarse significantly outperforms fine (row 4, note the negative effect size), while for B-factor, fine does not significantly outperform volume-alone (row 24). For all other comparisons, the improvement is modest but significant, supporting the efficacy of \modelname-derived features.}
\label{tab:per-volume-property-prediction-results}
\resizebox{\textwidth}{!}{%
\begin{tabular}{c l l l l c l}
\toprule
 & Task & Comparison & $p$ (Wilcoxon) & $p$ (Holm--Bonferroni) & Win rate & Hodges--Lehmann Effect (95\% CI) \\
\midrule
1  & sec.\ struc.  & fine > coarse        & $2.89\times10^{-13}$ (****)  & $2.31\times10^{-12}$ (****) & 0.642 & $+0.0092$, [0.0069, 0.0115] \\
2  & sec.\ struc.  & fine > random        & $2.13\times10^{-80}$ (****)  & $3.41\times10^{-79}$ (****) & 0.869 & $+0.0366$, [0.0343, 0.0389] \\
3  & sec.\ struc.  & fine > volume alone  & $1.78\times10^{-102}$ (****) & $3.56\times10^{-101}$ (****)& 0.938 & $+0.0612$, [0.0575, 0.0651] \\
4  & amino         & fine > coarse        & $3.78\times10^{-56}$ (****)  & $4.54\times10^{-55}$ (****) & 0.212 & $-0.0123$, [$-$0.0138, $-$0.0109] \\
5  & amino         & fine > random        & $5.73\times10^{-05}$ (****)  & $1.72\times10^{-04}$ (***)  & 0.527 & $+0.0026$, [0.0013, 0.0040] \\
6  & amino         & fine > volume alone  & $1.31\times10^{-100}$ (****) & $2.50\times10^{-99}$ (****) & 0.954 & $+0.0267$, [0.0255, 0.0279] \\
7  & mol.\ type    & fine > coarse        & $1.50\times10^{-07}$ (****)  & $8.99\times10^{-07}$ (****) & 0.599 & $+0.0029$, [0.0018, 0.0039] \\
8  & mol.\ type    & fine > random        & $4.96\times10^{-58}$ (****)  & $6.44\times10^{-57}$ (****) & 0.761 & $+0.0130$, [0.0116, 0.0146] \\
9  & mol.\ type    & fine > volume alone  & $2.96\times10^{-113}$ (****) & $6.52\times10^{-112}$ (****)& 0.974 & $+0.0572$, [0.0542, 0.0603] \\
10 & mol.\ class.  & fine > coarse        & $4.01\times10^{-07}$ (****)  & $2.01\times10^{-06}$ (****) & 0.587 & $+0.0032$, [0.0018, 0.0051] \\
11 & mol.\ class.  & fine > random        & $7.23\times10^{-24}$ (****)  & $6.50\times10^{-23}$ (****) & 0.708 & $+0.0076$, [0.0059, 0.0095] \\
12 & mol.\ class.  & fine > volume alone  & $4.09\times10^{-82}$ (****)  & $6.95\times10^{-81}$ (****) & 0.906 & $+0.0835$, [0.0694, 0.0973] \\
13 & hydro.        & fine > coarse        & $1.04\times10^{-12}$ (****)  & $7.26\times10^{-12}$ (****) & 0.626 & $+0.0049$, [0.0036, 0.0063] \\
14 & hydro.        & fine > random        & $8.46\times10^{-32}$ (****)  & $8.46\times10^{-31}$ (****) & 0.723 & $+0.0081$, [0.0069, 0.0093] \\
15 & hydro.        & fine > volume alone  & $1.10\times10^{-109}$ (****) & $2.31\times10^{-108}$ (****)& 0.980 & $+0.0471$, [0.0450, 0.0492] \\
16 & abs.\ SASA    & fine > coarse        & $1.46\times10^{-73}$ (****)  & $2.19\times10^{-72}$ (****) & 0.839 & $+0.0084$, [0.0076, 0.0091] \\
17 & abs.\ SASA    & fine > random        & $2.44\times10^{-72}$ (****)  & $3.42\times10^{-71}$ (****) & 0.821 & $+0.0065$, [0.0060, 0.0071] \\
18 & abs.\ SASA    & fine > volume alone  & $1.83\times10^{-114}$ (****) & $4.38\times10^{-113}$ (****)& 0.997 & $+0.0809$, [0.0787, 0.0832] \\
19 & rel.\ SASA    & fine > coarse        & $3.80\times10^{-99}$ (****)  & $6.84\times10^{-98}$ (****) & 0.929 & $+0.0223$, [0.0211, 0.0234] \\
20 & rel.\ SASA    & fine > random        & $2.74\times10^{-39}$ (****)  & $3.01\times10^{-38}$ (****) & 0.762 & $+0.0069$, [0.0060, 0.0077] \\
21 & rel.\ SASA    & fine > volume alone  & $1.94\times10^{-114}$ (****) & $4.46\times10^{-113}$ (****)& 0.996 & $+0.0850$, [0.0822, 0.0879] \\
22 & B-fact.       & fine > coarse        & $1.37\times10^{-03}$ (**)    & $2.74\times10^{-03}$ (**)   & 0.551 & $+0.0210$, [0.0081, 0.0344] \\
23 & B-fact.       & fine > random        & $9.86\times10^{-06}$ (****)  & $3.95\times10^{-05}$ (****) & 0.572 & $+0.0296$, [0.0164, 0.0436] \\
24 & B-fact.       & fine > volume alone  & $4.01\times10^{-01}$ (n.s.)  & $4.01\times10^{-01}$ (n.s.) & 0.490 & $+0.0062$, [$-$0.0081, 0.0210] \\
\bottomrule
\end{tabular}%
}
\end{table}

\section{Limitations and Future Work}
\label{sec:limitations}
We note several boundaries of our current evaluation that offer clear directions for future work.

\paragraph{Generalization to non-redundant structures.} Our train, validation, and test splits are random over EMDB IDs, following prior cryoEM benchmarks~\citep{cryo2structdata, emnuss}. All models in our comparison are trained and evaluated on the same splits, so the relative improvements in Table~\ref{tab:property-prediction-results} are not affected. However, random splitting can place homologous proteins or alternate conformations on both sides, which limits our ability to claim absolute generalization to entirely unseen folds. Re-evaluation on splits stratified by sequence identity or CATH topology is a natural next step.

\paragraph{Pretraining cost and high-frequency representation.} Pretraining \modelname{} requires roughly 720 H100-hours. Once trained, the model is fully amortized: a single forward pass produces $\theta_V$ for an unseen volume, and feature extraction is a deterministic offline precomputation. The pretraining cost remains a barrier to iteration and to scaling the framework to larger volumes or higher-resolution grids, and reducing it through better tokenization, sparser attention, or distillation is a promising direction. Separately, the highest frequencies in our Fourier coordinate encoding can exceed the Nyquist limit of the $96^3$ grid; the spectral bias of ReLU MLPs acts as an implicit regularizer in practice, but principled bandlimiting is a logical refinement.

\end{document}